Investigating Relative Performance of Transfer and Meta Learning

Benji Alwis
Bootham School
York
UK

**Abstract**

Over the past decade, the field of machine learning has experienced remarkable advancements. While image recognition systems have achieved impressive levels of accuracy, they continue to rely on extensive training datasets. Additionally, a significant challenge has emerged in the form of poor out-of-distribution performance, which necessitates retraining neural networks when they encounter conditions that deviate from their training data. This limitation has notably contributed to the slow progress in self-driving car technology. These pressing issues have sparked considerable interest in methods that enable neural networks to learn effectively from limited data.

This paper presents the outcomes of an extensive investigation designed to compare two distinct approaches, transfer learning and meta learning, as potential solutions to this problem. The overarching objective was to establish a robust criterion for selecting the most suitable method in diverse machine learning scenarios. Building upon prior research, I expanded the comparative analysis by introducing a new meta learning method into the investigation. Subsequently, I assessed whether the findings remained consistent under varying conditions. Finally, I delved into the impact of altering the size of the training dataset on the relative performance of these methods. This comprehensive exploration has yielded insights into the conditions favoring each approach, thereby facilitating the development of a criterion for selecting the most appropriate method in any given situation.

**1.Introduction**

Traditionally, machine learning algorithms have heavily relied on extensive datasets. However, several factors can hinder the application of neural networks when dealing with limited data. For instance, certain domains, like medical diagnostics, face challenges due to the rarity of specific conditions or privacy concerns. Additionally, practical constraints, such as ethical and safety considerations, can limit data availability, as seen in the case of drug discovery, where assessing the efficacy of new drug candidates can be hindered by the potential toxicity, low activity, and low solubility of these molecules. Furthermore, data collection can be prohibitively expensive in certain scenarios, such as space applications. Lastly, the scarcity of computational or financial resources can hinder the retraining of algorithms with large datasets.

In contrast, humans exhibit the remarkable ability to learn from just a few examples. A child, for example, can recognize a giraffe after seeing only a handful of photos. This quick recognition stems partly from the child's prior exposure to various living beings, which has made them familiar with the key features of animals. Few-shot learning aims to emulate this capability by leveraging common representations to learn from a small amount of training data. In this paper, I explore different approaches to achieve representative learning, with a particular focus on transfer learning and meta-learning.

Few-shot learning, at its core, is the capacity to learn from a limited number of examples. To compensate for the absence of extensive datasets, as is typically the case in traditional neural network learning, few-shot learning methods seek to exploit alternative sources of information. This additional information can either be pre-existing knowledge, as demonstrated in transfer learning (Weiss), or extra knowledge derived from the small dataset itself, as exemplified by meta-learning (Finn).

While both meta-learning and transfer learning have garnered increasing interest in the machine learning community, there is a notable gap in the literature regarding a comprehensive comparative analysis of these two approaches. This void exists because these approaches have evolved independently, and until now, there has been insufficient evidence to conclusively compare their effectiveness. Despite the inherent flexibility that meta-learning offers, an existing comparative study (Dumoulin) has, thus far, suggested that transfer learning outperforms meta-learning. However, it is crucial to note that this field of research is dynamic and continually evolving.

In this paper, my primary objective is to bridge this gap by conducting an in-depth investigation into both meta-learning and transfer learning approaches. My aim is to establish a criterion for selecting the most suitable method in a given machine learning scenario. To accomplish this, I have expanded upon previous research by incorporating a recently acclaimed meta-learning method into our comparative analysis.

My investigation involved subjecting both meta-learning and transfer learning methods to various experimental conditions known to influence the performance of machine learning algorithms. These rigorous experiments allowed me to draw meaningful conclusions regarding the relative strengths and weaknesses of these approaches. As a result, I have identified a robust criterion for selecting the most appropriate algorithm in specific real-world scenarios.

The remainder of this paper is structured as follows: Following an introduction to the primary objectives of this investigation, I delve into the concept of few-shot learning. Specifically, I explore two distinct strategies for few-shot learning, namely transfer learning and meta learning. These sections serve as the foundation for the experiments detailed in the subsequent section. Lastly, I offer a concise summary of the key conclusions drawn from this research.

## 2. Main Objectives

The primary objective of this investigation was to establish a comprehensive criterion for selecting the optimal few-shot learning method tailored to specific scenarios. Traditionally, the fields of transfer learning and meta learning have evolved independently. In pursuit of this aim, I conducted an extensive review of the current state-of-the-art in both of these learning methodologies, primarily drawing from published research literature. Subsequently, I undertook experimental comparisons of these two methods within a framework initially proposed by Dumoulin and colleagues, expanding upon their work by incorporating a newly developed meta learning approach. This addition was driven by the overarching question: "Do the prior conclusions regarding the relative performance of these two methods still hold true in light of recent advancements in meta learning?"

Furthermore, I addressed a critical question that remained unanswered in Dumoulin's work: How does the relative performance of these methods evolve when the size of the training dataset is altered? To explore this variation, I conducted an experiment, the findings of which have contributed to the formulation of a criterion for choosing the most suitable few-shot learning method in a given situation.

## 3. Two Alternatives for Few-shot Learning

Few-shot learning is characterised by the ability to acquire knowledge and make predictions from a limited set of examples (Wang). In contrast to traditional neural network learning, which often relies on abundant datasets, few-shot learning methods seek to harness alternative sources of information. This supplementary information can take the form of pre-existing knowledge, as seen in the field of transfer learning, or additional insights derived from the small dataset itself, a hallmark of meta-learning.

The diverse landscape of few-shot learning algorithms can be broadly categorised according to the elements they leverage to achieve this goal. Within a typical machine learning system, three fundamental components exist: the model or hypothesis space, the data, and the algorithm. Various few-shot algorithms are strategically designed to exploit one of these components, often in conjunction with prior knowledge, to enhance their learning capabilities.

### 3.1 Transfer Learning

The human visual perception system undergoes continuous refinement through prolonged exposure to visual stimuli and interactions with the environment. As a result, it becomes proficient in recognizing objects using a relatively small set of distinctive features. Central to the concept of transfer learning is representation learning, which focuses on acquiring and mastering the essential representative features necessary for automated visual recognition.

Transfer learning methods typically employ a two-step procedure. Initially, a neural network model undergoes pre-training using a sizable dataset referred to as the "upstream dataset."

Prominent examples of these upstream datasets include ImageNet and CIFAR. Subsequently, the model is fine-tuned using a task-specific dataset known as the "downstream dataset," which is comparatively smaller in size compared to the upstream dataset.

The fascination with transfer learning methods gained traction alongside the rapid advancements in Convolutional Neural Networks (CNNs) after 2012 (Albawi). These CNNs were originally trained on extensive collections of images, notably the ImageNet dataset. The most successful architectures, such as VGG (Simonyan) and ResNet (He), featured deep structures with a multitude of parameters that required optimization with substantial datasets. Initially, these models excelled in applications where ample data was available. However, adapting them to different applications became problematic when relevant image collections were limited or absent. Additionally, in cases where data was available, the computational resources required for retraining these models from scratch posed challenges.

Transfer learning methods emerged as a viable solution to these issues. Through transfer learning, a model initially trained on a typically large dataset can be fine-tuned using a smaller, task-specific dataset. For example, a neural network pre-trained on ImageNet can be fine-tuned to recognize various cat breeds, or a sentiment analyzer pre-trained on extensive movie review data can be fine-tuned for assessing sentiment in consumer goods reviews. In these applications, both pre-training and fine-tuning datasets were labelled, enabling both phases to follow the supervised learning paradigm. However, in the field of natural language processing (NLP), where labelled data is often scarce, transfer learning methods have been adapted to pre-train neural networks using unlabeled data (unsupervised learning) and fine-tune them with labelled data (supervised learning).

During the fine-tuning phase, not all pre-trained weights remain fixed; a subset of them is adjusted based on the new data. However, it is crucial to ensure that the number of parameters being adjusted does not lead to overfitting, especially when the size of the new dataset is significantly smaller than the number of available parameters. This risk can be mitigated by freezing part of the network. The choice of which layers to freeze is typically made based on heuristics (Kolesnikov). In convolutional neural networks, it has been observed that layers closer to the input handle generic image processing tasks, such as edge detection, while layers nearer to the output are more specialised for the target application (Xai).

Fine-tuning is often considered more of an art than a science (Stanford, Lecture 3), involving several critical decisions:

**3.1.1 Determining how to unfreeze the network layers**

During the initial stages of the fine-tuning process, the entire neural network is held in a frozen state. This entails that all network layers are equipped with pre-trained weights. As the fine-tuning progresses, the number of layers that are allowed to adapt to the new task can be incrementally increased. This is achieved by gradually unfreezing layers that are located further away from the output layer. The feasibility of this approach depends on the size of the task-specific dataset; a larger dataset can support the fine-tuning of a greater number of parameters. In essence, if the downstream dataset is substantial enough, it becomes plausible to fine-tune the entire network.

However, when dealing with a smaller downstream dataset, unfreezing a large number of layers is not advisable. This is because it can lead to overfitting, as discussed in the previous section. In such situations, it becomes essential to consider the similarity between the source and target datasets. If the two datasets exhibit a high degree of similarity, it may be viable to directly apply pre-trained weights to all network layers during fine-tuning. Conversely, if the datasets are markedly dissimilar, it could prove advantageous to discard the pre-trained weights in some of the later layers and initialise them with random weights. This is because the original weights may lack relevance to the target task due to the substantial dissimilarity between the datasets.

### 3.1.2 Determining the learning rate

Within a neural network, the learning rate plays a pivotal role in controlling the pace of learning, specifically how rapidly the weights are adjusted during training. A common and effective practice is to initiate training with a smaller learning rate, especially when dealing with the initial layers of the network.

### 3.1.3 The Best Available Method for Transfer Learning: Big Transfer (BiT) Method

The Big Transfer (BiT) method (Kolesnikov) has gained widespread recognition as one of the main approaches in the domain of transfer learning. It offers a relatively straightforward framework for transferring pre-trained general representations into specific problem domains or tasks. Much like other transfer learning methodologies, BiT leverages a vast dataset to acquire a comprehensive understanding of feature representations within the data. To illustrate, in the context of image processing, BiT has developed three distinct versions of its system, each trained on datasets of varying sizes. The largest among them has been pre-trained on an extensive dataset known as JFT-300M, consisting of a staggering 300 million noisily labelled images. This pre-training phase does demand substantial computational resources and time investment; however, it is a one-time expense, and the subsequent fine-tuning for downstream tasks incurs relatively lower costs.

BiT distinguishes itself from prior methods in two key ways. Firstly, it simplifies the fine-tuning process by relying on straightforward heuristics to adjust hyperparameters for specific tasks. This approach streamlines the fine-tuning step, making it more accessible and efficient. Secondly, BiT deviates from the common practice of employing Batch Normalisation (BN) for data normalisation during training. Instead, it adopts Group Normalisation (GN), based on their observation that BN performs inadequately and imposes significant computational overhead when used for pre-training in distributed computing environments. GN proves to be a more effective choice in this context.

In image recognition tasks, BiT predominantly employs the Residual Network (ResNet), a well-established neural network architecture.

Transfer learning remains an active area of research within the field of machine learning. Reported performance outcomes for transfer learning methods vary widely and are contingent on both the effectiveness of the algorithm and the datasets used in experimental settings.

### 3.1.4 Testing the Performance of Transfer Learning Methods

Until recently, the field of transfer learning lacked a universally accepted testing methodology, making it challenging to effectively compare different transfer learning methods. This difficulty stemmed from the fact that experiments often utilised distinct upstream (pre-training) and downstream (fine-tuning and testing) datasets.

However, a significant development occurred with the release of the Visual Task Adaptation Benchmark (VTAB) paper in 2020 (Zhai). VTAB quickly gained widespread recognition for providing a robust framework that is both demanding and practical, facilitating the assessment of new transfer learning methods.

As previously discussed, transfer learning entails the process of fine-tuning a pre-trained model using new datasets, referred to as downstream datasets. Each of these datasets is designed with a specific objective in mind, known as a downstream task. The formulation and selection of these downstream tasks represent a critical aspect of an evaluation benchmark, as they must offer a sound basis for comparative analysis.

Given VTAB's focus on visual recognition task evaluation, the benchmark is constrained to tasks that rely solely on visual inputs. Within VTAB, the chosen 19 tasks are categorised into three main groups: natural, specialised, and structured. Natural tasks encompass images captured using standard cameras and span a range from generic to fine-grained and abstract images. Specialised tasks involve images of the world captured using specialised cameras, including medical and remote sensing images. Finally, structured tasks pertain to scenes' structural attributes and can be employed for tasks like 3D depth prediction or object counting. These structured tasks are often constructed using simulated environments.

Importantly, VTAB does not impose constraints on upstream training or transfer strategies, allowing researchers the flexibility to experiment and explore different approaches. In actual VTAB experiments, researchers have limited the size of the downstream dataset to 1000 instances, simulating conditions akin to few-shot learning. Consequently, for each downstream task, a random subset of 1000 images is selected for fine-tuning the neural network model. The assessment of performance in each task is conducted using another non-overlapping subset of 600 random images (referred to as the query dataset) from the same dataset collection. Performance is quantified as the average classification accuracy over the entire query set, providing a comprehensive evaluation of transfer learning methods.

## 3.2 Meta Learning

Traditional deep learning faces a significant limitation: once a neural network is trained for a specific task, it cannot be readily repurposed for a closely related task without undergoing complete retraining from scratch. This issue is precisely what meta learning aims to address. It does so by cultivating a shared set of parameters among a group of interconnected tasks, enabling the model's reuse for new, related tasks. Meta learning offers a compelling advantage: it can accomplish tasks with minimal pre-existing data. An illustrative example is training an AI system to mimic the writing style of a particular individual or to adapt to a different language based on prior knowledge. Meta learning provides a viable alternative in scenarios where collecting extensive training examples is challenging due to data scarcity or resource constraints. Additional instances of its potential include per-language handwritten character recognition and personalised spam filtering. A common feature among these tasks is the presence of numerous tasks (e.g., languages for character recognition or individuals for personalised spam preferences), each with relatively limited examples. In standard deep learning approaches, success often hinges on the availability of abundant examples. However, in the context of meta learning, each language or individual in the mentioned examples is considered a distinct task. For instance, in handwritten character recognition, there are shared characteristics within each language, and for spam recognition, there are common instances of spam applicable to almost everyone. Meta learning strives to separately discern these shared traits and task-specific attributes, facilitating adaptation to new tasks using a comparatively small dataset, with additional support from the shared characteristics.

Meta learning algorithms maintain two sets of parameters: one set is specific to individual tasks, while the other set is universal across all tasks. The latter is termed "meta parameters" and is determined through learning across a range of related tasks. In some research literature, the concept of learning to learn is described as the ability to choose a learning algorithm based on input data characteristics []. However, this notion is more aligned with automated machine learning rather than pure meta learning (Yaliang).

In a parallel manner to how neural network training seeks to enhance generalisation abilities (i.e., accuracy on unseen data), meta learning aims to elevate task generalisation (i.e., accuracy on unseen tasks). The concepts of meta-overfitting and meta-underfitting are analogous to overfitting and underfitting in traditional neural networks. For instance, meta-overfitting occurs when parameters are excessively optimised for the training tasks (Stanford, Lecture 4).

There are two prevailing viewpoints in the area of meta learning: the mechanistic view and the probabilistic view. In the mechanistic view, meta learning is essentially the training of a neural network using a meta-dataset, which encompasses multiple datasets, each associated with a different task. In contrast, the probabilistic view frames meta learning as the extraction of prior information from a set of tasks, enabling the efficient learning of new tasks. This prior knowledge is then leveraged to learn a new task with a limited number of training examples by inferring the most likely parameters.

### 3.2.1 Categorisation of Meta Learning Methods

Meta learning algorithms can be broadly categorised into four groups (Huisman).

- metric-based (or non-parametric) methods
- initialisation-based methods
- optimization-based methods
- model-based (or black box) methods

**Metric-based meta learning methods** are designed to acquire an appropriate metric space for the task at hand. For instance, when the goal is to discern the similarity between two images, a straightforward neural network is employed to extract features from both images. Subsequently, the similarity is quantified by measuring the distance between the features of these two images. The fundamental concept here revolves around task-specific learning, achieved by contrasting validation points with training points to predict the labels of corresponding training points. This approach finds extensive application in few-shot learning scenarios, where data points are limited in number. Noteworthy examples of metric-based learning algorithms encompass siamese networks (Melekhov), prototypical networks (Snell), graph neural networks (Zhou), and relation networks (Hu).

**Initialization-based meta learning** methods are primarily focused on the acquisition of optimal initial parameter values. In a typical neural network training pipeline, random weights are initially assigned to the network. Subsequently, the process involves computing the loss by comparing the generated outputs with the expected outputs and subsequently minimising this loss through an optimization method, often gradient descent. The overarching goal is to calculate optimal weights with the objective of minimising errors. However, initialising the network with random weights can introduce challenges, making training more arduous and necessitating a substantial amount of data to achieve a satisfactory level of accuracy. This becomes particularly problematic when training data availability is limited. This challenge can be surmounted by initialising weights with values that are in close proximity to the optimal values, thereby expediting neural network convergence and learning. Initialization-based meta learning methods strive to compute these optimal weights through the learning of meta parameters. Notable algorithms in this category include MAML (Liu), Reptile (Nichol), and Meta-SGD (Li).

During meta-learning, the full set of parameters is divided into two distinct groups. Meta parameters, also known as outer parameters, encompass those that are learned at the meta level. In contrast, task-specific parameters, often referred to as inner parameters, pertain to those fine-tuned for specific tasks. Achieving this segregation, however, has posed challenges as it necessitates identifying a subset of parameters suitable for meta-learning. Additionally, questions have arisen regarding whether a single initial condition suffices when confronted with a range of diverse tasks. Consequently, meta-learning algorithms have emerged that employ combinations of initial conditions to address this concern.

**Optimization-based methods** in the field of few-shot learning place a distinct emphasis on the acquisition of an optimised learning strategy. This emphasis arises from the inherent challenge of working with a limited amount of data in few-shot learning scenarios, which often proves insufficient for minimising errors through conventional optimization techniques like gradient descent. In these approaches, a two-network architecture is typically employed: a base network tasked with actual learning, and a meta network responsible for optimising the base network. Such methods find applicability in few-shot learning and in the enhancement and acceleration of many-shot learning.

**Model-based methods** represent a strategic departure from the conventional practice of iteratively computing optimal weights when dealing with limited datasets. Instead of this iterative approach, these methods employ a model to directly derive the optimal set of weights. This family of techniques is often referred to as black box models, owing to their relative lack of transparency and interpretable mechanisms.

### 3.2.2 Key Challenges in Meta Learning

Meta learning confronts two main challenges, the first of which is known as "negative transfer." This occurs when meta learning fails to provide a substantial advantage, making learning through independent networks a more favorable choice. Negative transfer typically arises due to two primary contributing factors. The first factor stems from optimization difficulties triggered by the varying learning rates across different tasks or parameter interference between tasks, a phenomenon referred to as parameter mixup. The second contributing factor relates to the limited representation capacity of the network. This limitation arises from the necessity for meta-learning networks to be larger in size compared to traditional neural networks, as they must learn both meta and task-specific parameters simultaneously.

### 3.2.3 Example Meta Learning Methods

The following meta learning methods were used in the experiments discussed in section 4.

**Prototypical Networks** is a method that falls into the category of metric based meta learning methods.

**ProtoMAML** is a variant of the MAML that falls into the category of initialisation based meta learning methods.

**Data2Vec** (Baevski) is a method that falls into the category of metric based meta learning methods explained in section 3.3.1.

### 3.2.4 Testing Meta Learning Algorithms for Few-shot Learning

Experimental Procedure

A standard neural network is tested for its ability to correctly classify a previously unseen example. In few-shot learning, the ability to correctly classify a previously unseen example using a neural network trained using only a small dataset is tested. The following key steps are followed in meta learning training and subsequent testing.

- Select a dataset
- Randomly choose a subsample of classes
- Randomly choose examples within the previously chosen subsample of classes and assign them either to the training set (support set) or the query set
- Conduct training and subsequent testing using the constructed support and query sets.

In meta learning, this approach to training is called episodic learning due to the way training episodes are constructed by random selection. Figure 1 shows an example meta learning training and test phase. In this example, each row in the meta-training block represents a training episode. In each episode, five classes have been randomly chosen and one example image per each class has also been chosen. The same approach has been used in constructing test episodes.

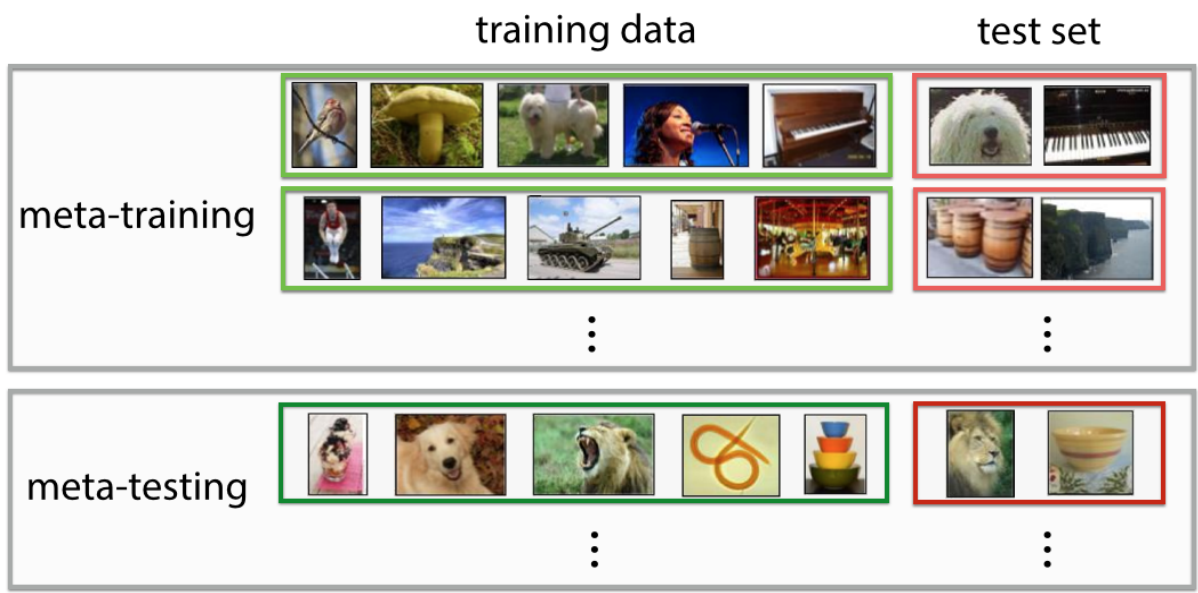

Figure 1 - An example meta learning process. (image credit - Stanford Meta Learning Lecture Series).

**Datasets**

Until recently the two main datasets used for meta learning experiments were the Omniglot and mini-ImageNet. The Omniglot dataset contains 1623 handwritten characters taken from 50 different alphabets. Each data class contains 20 characters. Mini-ImageNet is a subset of the standard ImageNet dataset. It contains 100 ImageNet classes. Each class contains 600 examples. However, it has been recently argued that these datasets are not challenging or realistic enough to show improvements of the new methods and pave the way for further advancement. The other limitations highlighted include the homogeneity of the learning tasks (real life learning experiences are heterogeneous) and the measuring of generalisation ability only within the dataset.

The Meta-Dataset aims to overcome these limitations by introducing a realistic heterogeneity by varying the number of classes in each task and the size of the training set. It provides a more realistic environment to test few-shot learners. The size of the dataset is larger than the previously mentioned datasets and it is more diverse as well. This dataset has been constructed using 10 previous datasets. They are ILSVRC-2012, Omniglot, Aircraft, CUB-200-2011, Describable Textures, Quick Draw, Fungi , VGG Flower), Traffic Signs and MSCOCO. They contain images of different natural and human-made objects. There is a large variation of the granularity of images that increases the challenge of image recognition and generalisation.

## 4. Experiments

### 4.1 Main Objective

The primary objective of this investigation was to establish a robust criterion for selecting the most suitable few-shot learning method in specific scenarios.

In the initial phase of experiments, the primary goal was to compare the performance of transfer learning and meta learning methods within the context of few-shot learning problems. This choice was motivated by a study conducted by Dumoulin in early 2021, which notably favored transfer learning methods over meta learning methods when applied to few-shot learning challenges. Additionally, a newly developed meta learning method known as Data2Vec was included to explore whether it had made strides in narrowing the performance gap. Subsequently, in the second experiment, the evaluation was extended to assess whether these findings persisted when utilising higher quality images.

Following the observations from these initial experiments, the investigation delved into the impact of reducing dataset size. There have been claims suggesting that meta learning could yield impressive results with relatively small datasets (Stanford). However, comparative experiments that assessed the relative performance of transfer and meta learning as dataset size was progressively decreased were lacking. This served as the objective of the third experiment, a crucial step in developing a comprehensive criterion for choosing the optimal few-shot learning method tailored to specific circumstances.

## 4.2 Experimental Framework

### 4.2.1 The Computing Setup

Given that the core of the experiments revolved around the execution and evaluation of neural networks, it was important to establish the requisite hardware and software specifications.

To implement the neural networks, the TensorFlow software library (https://www.tensorflow.org/) was employed. TensorFlow is a prominent open-source software library that enjoys extensive usage in the field of neural network development. The experiments were conducted on a personal computer (PC) running the Linux operating system, featuring an Intel Core i7 processor, 16 GB of physical memory, and a NVIDIA GPU card. The GPU card played an important role in accelerating neural network training through parallelization, significantly enhancing computational efficiency.

### 4.2.1 The Dataset

Initially transfer learning and meta learning had been tested using different benchmarks. VTAB summarised in section 3.1.4 is one of the widely accepted benchmarks for transfer learning. Meta dataset , summarised in section 3.2.4, is one of the recent benchmarks for meta learning. In the previous study **(Dumoulin)**, these two benchmarks were combined to provide a common framework. It was called VTAB+MD. I used the same setup since the objective was to compare against their results.

In addition to the datasets, the Tensorflow hub makes pre-trained neural network models available. This helps save time and resources needed for training large neural networks from scratch. In this investigation, a pretrained neural network model was obtained from the Tensorflow Hub (https://www.tensorflow.org/hub). The particular model I used was called ResNet50 v2. It contained the parameters pre-trained using the large ImageNet dataset.

### 4.2.3 Experiments

The experimentation process was conducted in three distinct phases, each geared towards the overarching goal of pinpointing the most effective few-shot learning method tailored to specific situations.

In the first experiment, the primary aim was to address the question: "Can the novel meta-learning method, Data2Vec, bridge the performance gap between transfer learning and conventional meta learning methods?" This experiment closely followed the parameters employed by Dumoulin and colleagues in their prior experiments on transfer and meta learning methods.

With insights gained from the initial experiment, the focus shifted to the second phase. Here, the principal objective was to investigate the validity of the results when working with images of varying sizes, effectively representing different quality inputs. This inquiry was crucial because higher quality images have the potential to yield superior results, and the investigation sought to ascertain whether the same overarching trends persisted under these circumstances.

The third pivotal question sought to answer whether the supremacy of transfer learning over meta-learning remained intact when dealing with significantly reduced dataset sizes. This inquiry was motivated by the desire to explore whether the conclusions held when data availability was markedly limited.

The culmination of these inquiries was designed to provide a comprehensive answer to the ultimate question: "How can the most suitable few-shot learning method be selected in a given scenario?"

**Experiment 1**

Experimental Procedure

In total there were 29 datasets. Training and testing procedures were run independently for each dataset.

For each dataset, 1000 images were randomly chosen and assigned to the training set. Another non-overlapping subset of 600 images were randomly chosen and assigned to the query set.

The transfer learning method used was the Big Transfer (BiT) method together with the ResNet50 v2 model. It was fine tuned using the chosen 1000 images and tested individually against each of the chosen 600 images. This procedure was repeated 100 times. Correct number of classifications was expressed as a percentage and averaged for 5 categories of datasets: VTAB (All); VTAB (Natural); VTAB (Specialised); VTAB (Structured); and Meta Dataset.

To assess performance of meta learning methods, Prototypical Networks, ProtoMAML and Data2Vec were used. They were trained using the same training dataset of 1000 images and tested using the same dataset of 600 images. However, following the standard procedure for meta learning, as explained in section 3.3.2, the following additional steps were executed. Randomly choose a subsample of classes and then randomly choose examples within the previously chosen subsample of classes and use them as training or test episodes as shown in figure 1. Again this procedure was repeated 100 times. Correct number of classifications was expressed as a percentage and averaged for 5 categories of datasets: VTAB (All); VTAB (Natural); VTAB (Specialised); VTAB (Structured); and Meta Dataset. Appendix 1.2 shows part of the progress display during the training process.

Results were written into a file which was then automatically analysed for accuracy against the expected results. The overall performance was measured as the classification accuracy averaged over all test images. Results were aggregated for each data source.

Results

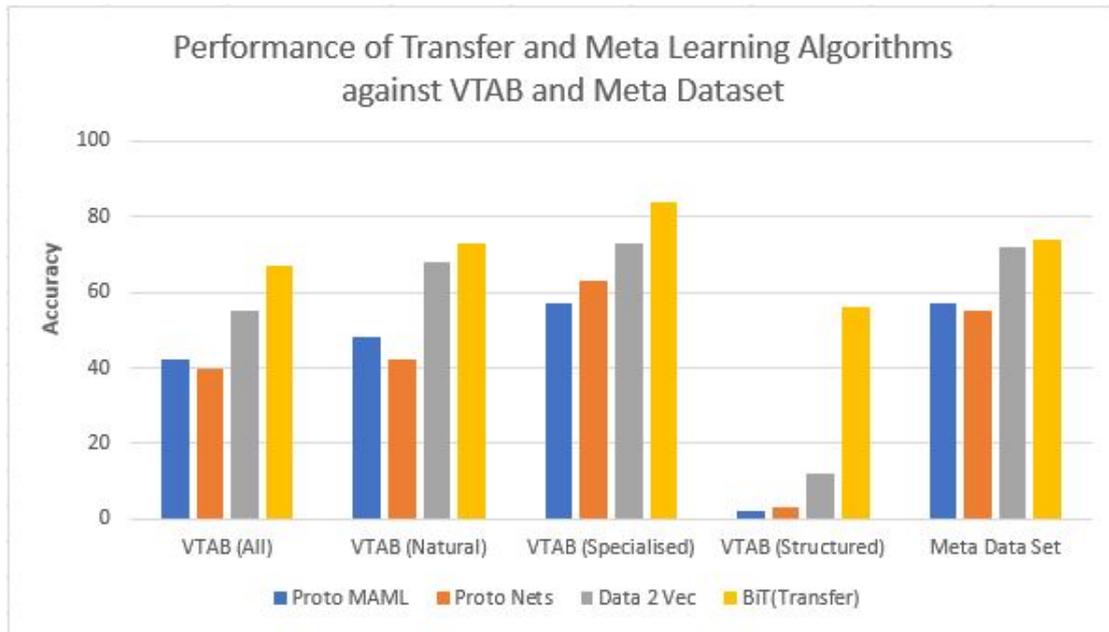

Figure 2 - Results obtained from running transfer and meta learning algorithms against VTAB and Meta Dataset.

Figure 2 shows the results obtained using different transfer and meta learning methods for different data sources. As previously known, the transfer learning method BiT, that was explained in section 3.1.3, performs better than the meta learning methods when tested against the data sources (natural, specialised and structured) from the transfer learning benchmark VTAB. This is not surprising since these data sources are more suited for transfer learning. It has been previously reported that even against the Meta-dataset which was aimed for assessing meta learning algorithms, Prototypical Networks and ProtoMAML under-perform in comparison to transfer learning. My results re-confirm this observation. However, I have found that this performance gap is reduced by relatively better performance of the new meta learning method Data2Vec which has not been previously tested in this context. For the Meta Dataset, Data2Vec performs almost the same level as the transfer learning method. This is encouraging since these results show that the state of art of meta learning methods have improved.

Another observation is that the meta learning methods perform better with the Meta dataset. One reason for this behaviour may be that the tasks within this dataset show higher degree of correlation than the tasks in the other datasets. This behaviour is consistent with the previous observations that meta learning methods work better with correlated datasets but under-perform in cross-dataset generalisation (Chen).

**Experiment 2**

Experimental Procedure

The main difference of this experiment was that input images with resolution of 224X224 were used. The resolution of the images used in the previous experiment was 126X126.

Results

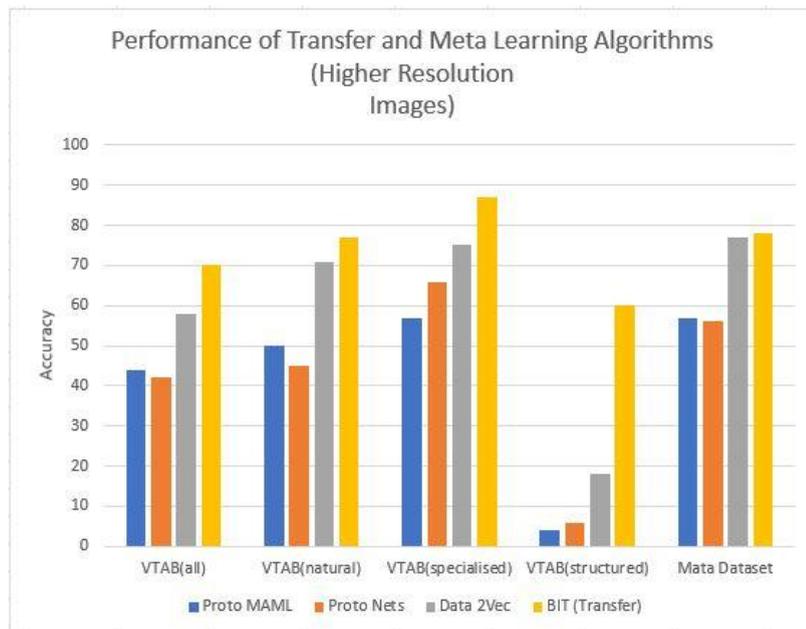

Figure 3 - Results obtained from running transfer and meta learning algorithms against VTAB and Meta Dataset using higher resolution images.

Figure 3 shows the results obtained using higher resolution images. It shows that all methods perform better with higher resolution images. However, the overall distribution of performance stays the same. This provides further evidence that the new meta learning methods, Data2Vec has managed to bridge the gap between transfer learning and meta learning methods even though transfer learning still outperforms meta learning under the experimental conditions used in these two experiments.

**Experiment 3**

Experimental Procedure

The main difference of this experiment was that experiment 1 was performed using 1000 training images and 600 test images. In experiment 3, the same experimental procedure was repeated with 750, 500, 250 and 100 training images. Accordingly, the test was also reduced to 450, 300, 150 and 60 respectively. Results were separately recorded.

**Results**

Figures 4 - 8 summarise the results obtained from the second experiment by incrementally reducing the size of the training dataset.

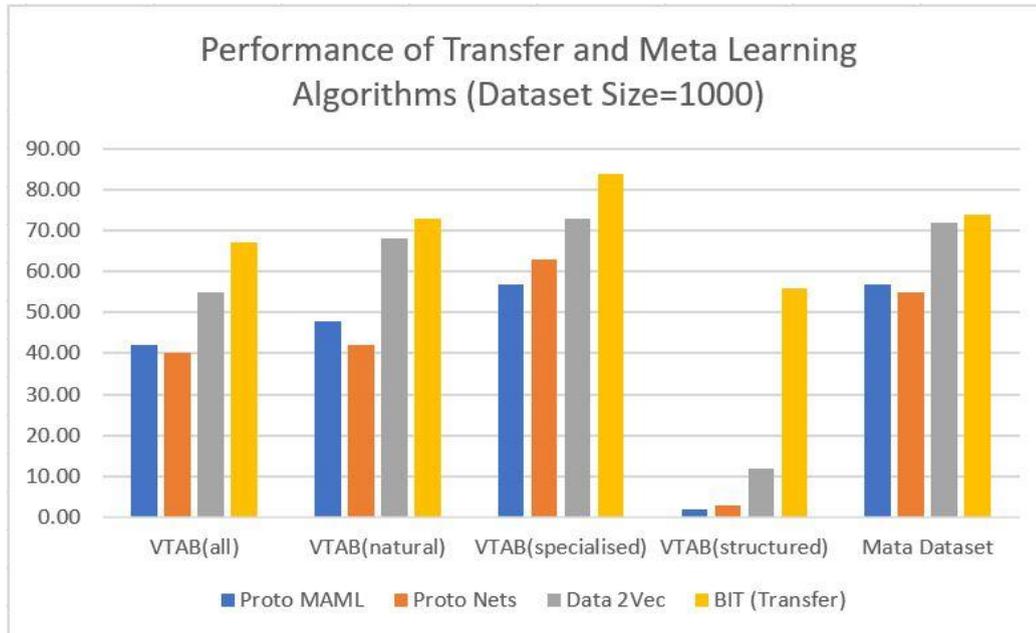

Figure **4** - Results obtained from running transfer and meta learning algorithms against VTAB and Meta Dataset when the size of the training dataset is 1000 images.

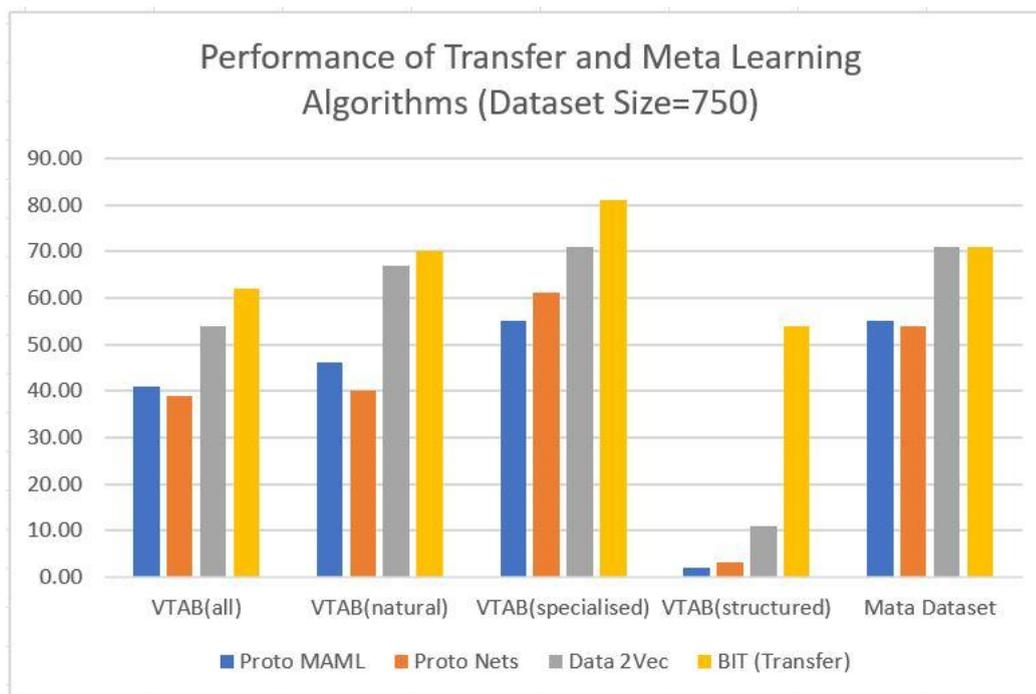

Figure **5** - Results obtained from running transfer and meta learning algorithms against VTAB and Meta Dataset when the size of the training dataset is 750 images.

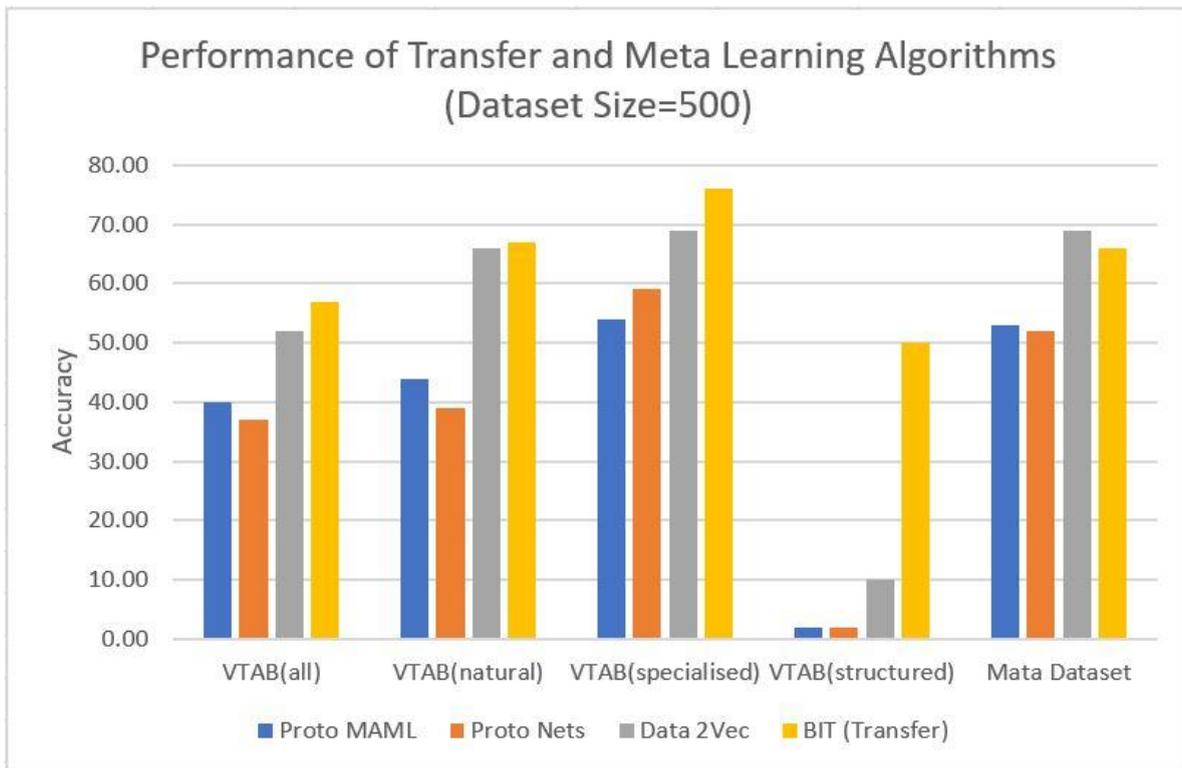

Figure **6** - Results obtained from running transfer and meta learning algorithms against VTAB and Meta Dataset when the size of the training dataset is 500 images.

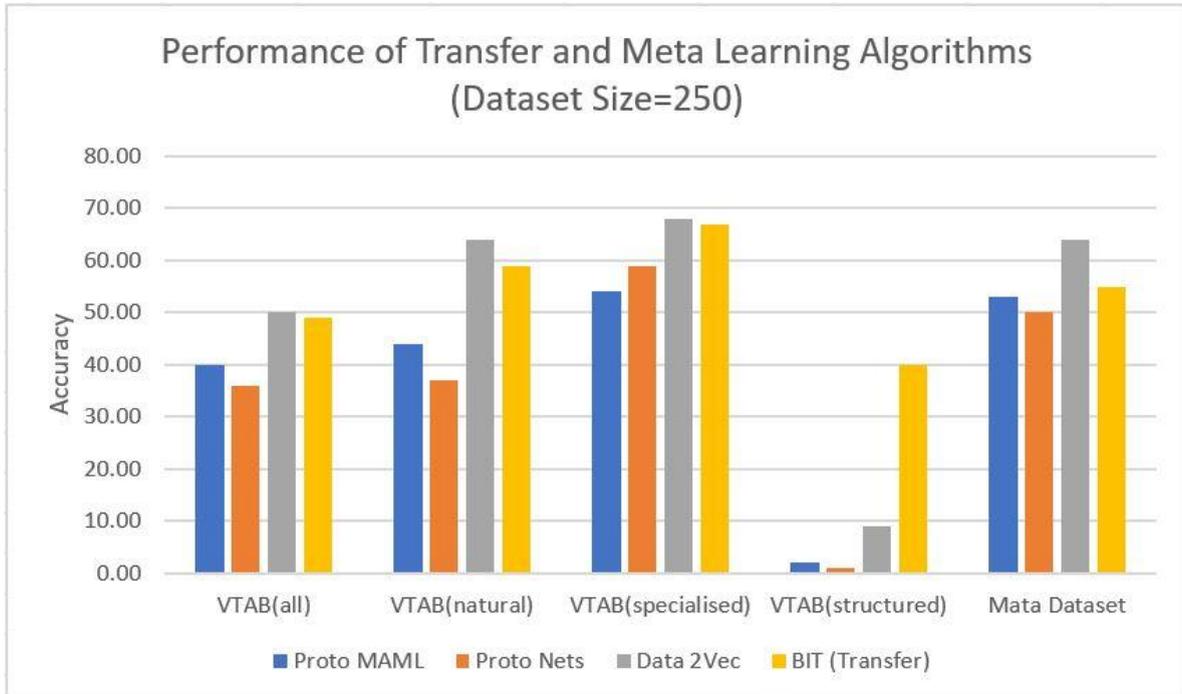

Figure **7** - Results obtained from running transfer and meta learning algorithms against VTAB and Meta Dataset when the size of the training dataset is 250 images.

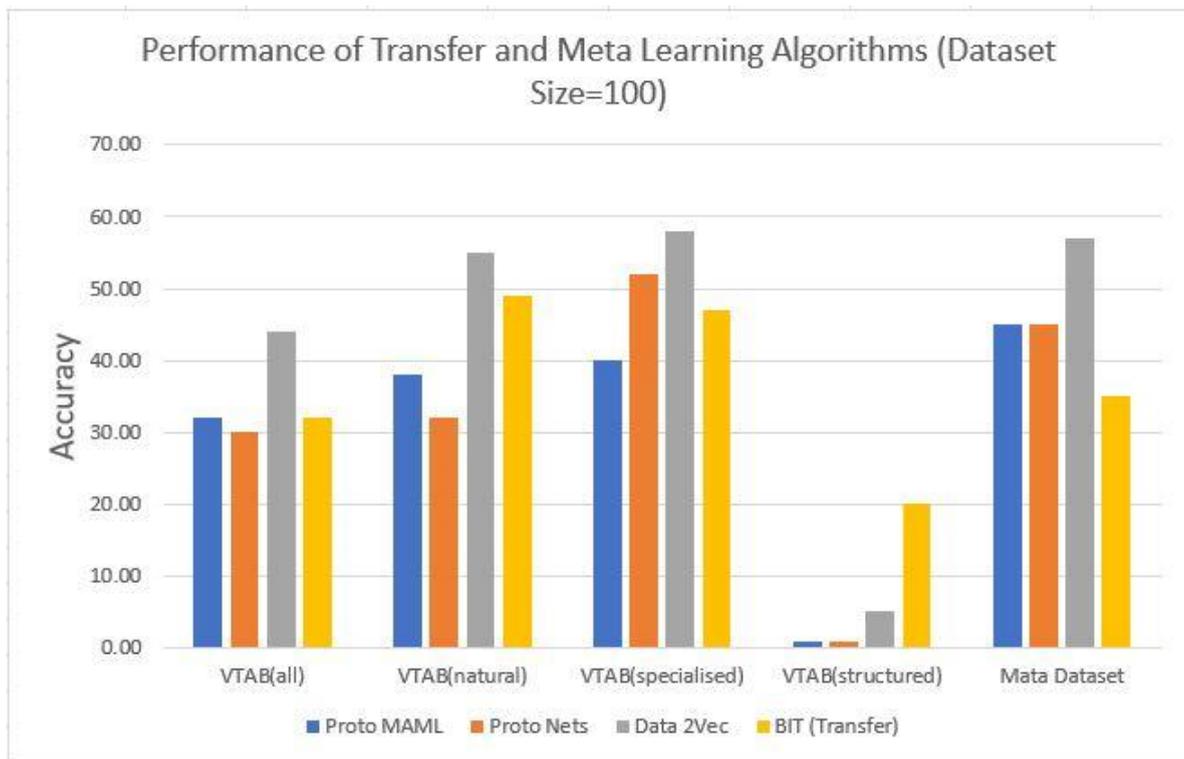

Figure **8** - Results obtained from running transfer and meta learning algorithms against VTAB and Meta Dataset when the size of the training dataset is 100 images.

The results from this experiment were very interesting. When the size of the training dataset was incrementally reduced down to 500, the overall superiority of the transfer method stood ground except against the Meta Dataset which was more suited for meta learning. However, when the size of the dataset was further reduced, meta learning methods, in particular the Data2Vec method outperformed the transfer method even though all methods suffered with the reduction of the size of the training dataset. These results showed that transfer learning methods are more affected by the reduction of the size of the training dataset and meta learning methods are more suited for few-shot learning when the size of the training dataset is very small.

## 5. Conclusions

Throughout the literature review, it became evident that the interest in few-shot learning has surged over the past five years. This surge in interest closely aligns with the burgeoning activity and performance improvements in transfer learning that began around 2015, driven by remarkable achievements in Convolutional Neural Networks leveraging vast datasets. It is conceivable that AI experts recognized the potential of transfer learning to replicate similar successes in domains where data scarcity prevailed. The enthusiasm for meta learning emerged slightly later, with a distinctive focus on achieving robust results for previously unencountered tasks characterised by limited data availability. Meta learning approaches, by their very nature, are more versatile than transfer learning, potentially offering higher utility.

Nevertheless, Dumoulin and colleagues have indicated that they have yet to match the performance of transfer learning methods. However, my results from experiment 1, involving the novel method Data2Vec from MetaAI, suggest that this performance gap is closing. Data2Vec was chosen for this experiment based on its impressive achievements in the area of natural language processing. While there was insufficient time for an in-depth analysis of the reasons behind its superior performance, further examination of this aspect promises valuable insights.

Perhaps the more noteworthy revelation pertains to how both methods respond when the training dataset size is reduced. Despite the prior findings, which suggested that transfer learning necessitates relatively large datasets for effective fine-tuning, my experiment indicated that they require a minimum of approximately 500 training samples. In scenarios where this requirement cannot be met, meta learning emerges as the superior approach.

These observations have contributed to the development of a criterion for selecting the most suitable few-shot learning method under varying circumstances, as graphically depicted in Figure 9. In conclusion, when the original dataset is available and there are ample computational resources and time, the optimal choice is to combine the two datasets. However, in practical scenarios where the original dataset is not accessible, the primary decision criterion becomes dataset size. If the dataset is sufficiently large for fine-tuning, transfer learning stands as the superior choice, provided there exists a meaningful correlation between the original and new tasks. Conversely, when dataset size falls short, meta learning emerges as the preferred alternative. It is important to note that meta learning heavily relies on task correlations, and in their absence, it yields notably subpar results. Thus, to summarise, for few-shot learning tasks, when dealing with a very small new dataset and substantial inter-task correlations, meta learning offers the superior approach.

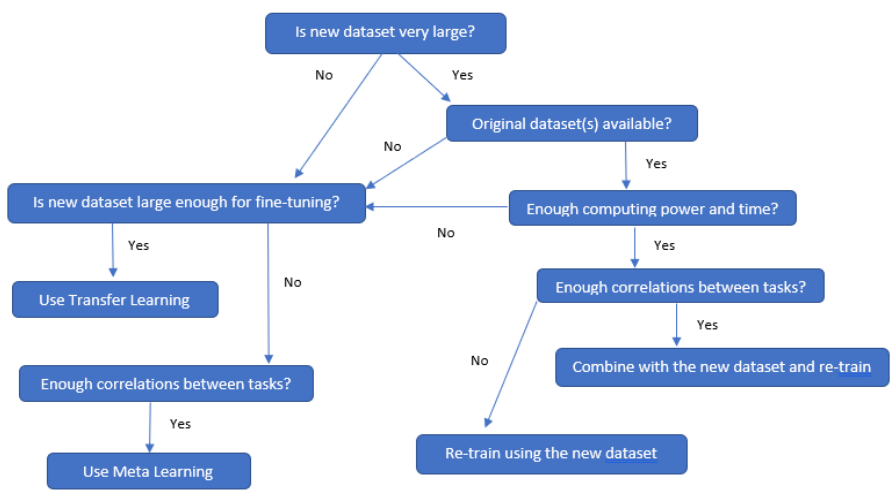

Figure 9 - The decision tree developed in this project to choose between transfer learning and meta learning.

**Bibliography**
1. Weiss, Karl, Taghi M. Khoshgoftaar, and DingDing Wang. "A survey of transfer learning." *Journal of Big data* 3.1 (2016): 1-40.
2. Gilmer, J., et al. "ICML'17: Proceedings of the 34th International Conference on Machine Learning." (2017).
3. Dumoulin, Vincent, et al. "A unified few-shot classification benchmark to compare transfer and meta learning approaches." *Thirty-fifth Conference on Neural Information Processing Systems Datasets and Benchmarks Track (Round 1)*. 2021.
4. Wang, Yaqing, et al. "Generalizing from a few examples: A survey on few-shot learning." *ACM computing surveys (csur)* 53.3 (2020): 1-34.
5. Albawi, Saad, Tareq Abed Mohammed, and Saad Al-Zawi. "Understanding of a convolutional neural network." *2017 international conference on engineering and technology (ICET)*. Ieee, 2017.
6. Simonyan, Karen, and Andrew Zisserman. "Very deep convolutional networks for large-scale image recognition." *arXiv preprint arXiv:1409.1556* (2014).
7. He, Kaiming, et al. "Deep residual learning for image recognition." *Proceedings of the IEEE conference on computer vision and pattern recognition*. 2016.
8. Stanford "Transfer & Meta-Learning." Deep Multi-task & Meta Learning ( 2020 ) Lecture 3. Stanford CS330, Feb. 2022.
9. Kolesnikov, Alexander, et al. "Big transfer (bit): General visual representation learning." *Computer Vision–ECCV 2020: 16th European Conference, Glasgow, UK, August 23–28, 2020, Proceedings, Part V 16*. Springer International Publishing, 2020.
10. Li, Yaliang, et al. "Automl: From methodology to application." *Proceedings of the 30th ACM International Conference on Information & Knowledge Management*. 2021.
11. "Transfer & Meta-Learning." Deep Multi-task & Meta Learning ( 2020 ) Lecture 4. Stanford CS330, Feb. 2022.
12. Huisman, Mike, Jan N. Van Rijn, and Aske Plaat. "A survey of deep meta-learning." *Artificial Intelligence Review* 54.6 (2021): 4483-4541.
13. Melekhov, Iaroslav, Juho Kannala, and Esa Rahtu. "Siamese network features for image matching." *2016 23rd international conference on pattern recognition (ICPR)*. IEEE, 2016.
14. Snell, Jake, Kevin Swersky, and Richard Zemel. "Prototypical networks for few-shot learning." *Advances in neural information processing systems* 30 (2017).
15. Zhou, Jie, et al. "Graph neural networks: A review of methods and applications." *AI open* 1 (2020): 57-81.
16. Hu, Han, et al. "Local relation networks for image recognition." *Proceedings of the IEEE/CVF International Conference on Computer Vision*. 2019.
17. Liu, Hao, Richard Socher, and Caiming Xiong. "Taming maml: Efficient unbiased meta-reinforcement learning." *International conference on machine learning*. PMLR, 2019.
18. Nichol, Alex, and John Schulman. "Reptile: a scalable meta learning algorithm." *arXiv preprint arXiv:1803.02999* 2.3 (2018): 4.
19. Li, Zhenguo, et al. "Meta-sgd: Learning to learn quickly for few-shot learning." *arXiv preprint arXiv:1707.09835* (2017).
20. Baevski, Alexei, et al. "Data2vec: A general framework for self-supervised learning in speech, vision and language." *International Conference on Machine Learning*. PMLR, 2022.
21. Chen, Wei-Yu, et al. "A closer look at few-shot classification." *arXiv preprint arXiv:1904.04232* (2019).
22. Zhai, Xiaohua, et al. "A large-scale study of representation learning with the visual task adaptation benchmark." *arXiv preprint arXiv:1910.04867* (2019).